\documentclass[letterpaper]{article} 
\usepackage[preprint]{aaai2027}
\usepackage[hyphens]{url}  
\usepackage{graphicx} 
\usepackage{natbib}  
\usepackage{caption} 
\usepackage{algorithm}
\usepackage{algorithmic}
\usepackage{amsmath,amssymb}

\usepackage{newfloat}
\usepackage{listings}
\DeclareCaptionStyle{ruled}{labelfont=normalfont,labelsep=colon,strut=off} 
\floatstyle{ruled}
\newfloat{listing}{tb}{lst}{}
\floatname{listing}{Listing}

\usepackage{booktabs}
\usepackage{microtype}
\usepackage{multirow}
\usepackage{graphicx}
\usepackage{booktabs} 
\usepackage{enumitem}
\usepackage[most]{tcolorbox}
\usepackage{pifont}
\usepackage{array}

\usepackage{tcolorbox}
\tcbuselibrary{skins}
\definecolor{ForestGreen}{rgb}{0.13,0.55,0.13}
\newtcolorbox{promptbox}[1][]{
  enhanced,
  colback=white,
  colframe=gray!70!black,
  coltitle=white,
  title=#1,
  fonttitle=\bfseries\large,
  boxrule=0.5pt,
  arc=2mm,
  left=1em, right=1em, top=1.6em, bottom=1em,
  attach boxed title to top left={xshift=0.5em,yshift=0em},
  boxed title style={
    colback=gray!70!black,
    arc=2mm,
    outer arc=0pt,
  },
  fontupper=\rmfamily,
  breakable,
  before=\vspace{1em}, 
  after=\vspace{1em},
  parbox=false, 
  width=\textwidth, 
  leftrule=1pt,
  rightrule=1pt, 
  toptitle=0mm,
  bottomtitle=0mm
}
\title{RoboChemGym: A Protocol-Driven Generative Simulation Framework for Long-Horizon Chemical Manipulation}

\author{
    Chenxi Li\textsuperscript{\rm 1,\rm 2}\equalcontrib,
    Haiyuan Wan\textsuperscript{\rm 3}\equalcontrib,
    Rui Li\textsuperscript{\rm 1,\rm 4},
    Jingyuan Li\textsuperscript{\rm 1,\rm 5},\\
    Sha Zhang\textsuperscript{\rm 6},
    Bohan Feng\textsuperscript{\rm 1,\rm 7},
    Jianbao Cao\textsuperscript{\rm 8},
    Zhangrui Zhao\textsuperscript{\rm 1,\rm 9},\\
    Di Hu\textsuperscript{\rm 10},
    Wangmeng Zuo\textsuperscript{\rm 4},
    Shixiang Tang\textsuperscript{\rm 1,\rm 6},
    Minting Pan\textsuperscript{\rm 1}\corresponding,
    Dongzhan Zhou\textsuperscript{\rm 1}\corresponding
}

\affiliations{
    \textsuperscript{\rm 1}Shanghai Artificial Intelligence Laboratory\\
    \textsuperscript{\rm 2}Zhejiang University\\
    \textsuperscript{\rm 3}Tsinghua University\\
    \textsuperscript{\rm 4}Harbin Institute of Technology\\
    \textsuperscript{\rm 5}Shanghai Jiao Tong University\\
    \textsuperscript{\rm 6}The Chinese University of Hong Kong\\
    \textsuperscript{\rm 7}Fudan University\\
    \textsuperscript{\rm 8}Wuhan University\\
    \textsuperscript{\rm 9}Beihang University\\
    \textsuperscript{\rm 10}Renmin University of China
}

\begin{document}

\maketitle

\begin{abstract}
Wet-lab experimentation serves as the gold standard for hypothesis verification in scientific discovery; yet it is inherently labor-intensive, costly, and safety-critical. Embodied agents hold the promise of automating these tedious workflows, but their development is hindered by the scarcity of real-world training data. While simulation offers a scalable alternative for producing demonstrations, current methods primarily target relatively short-horizon tasks with loosely structured interactions, failing to meet the strict procedural constraints and fine-grained manipulation demands of chemical experiments. To bridge this gap, we introduce \textbf{RoboChemGym}, a framework that autonomously generates high-fidelity manipulation demonstrations aligned with real-world experiment protocols, featuring a \textit{self-improving task synthesis} mechanism to iteratively refine task execution and scene configurations, enabling the reliable generation of expert trajectories for complex, multi-object protocols exceeding 10 interaction steps. Furthermore, we introduce a hierarchical benchmark that systematically assesses performance across varying granularities, spanning from atomic operations to full-cycle experimental workflows. RoboChemGym sets a scalable paradigm for the automated data synthesis and capability evaluation of embodied agents in intricate chemical tasks, serving as a critical stepping stone toward fully intelligent laboratories.
\end{abstract}


\section{Introduction}

Autonomous laboratories represent a transformative paradigm for scientific discovery, facilitating rapid hypothesis verification, standardized data collection, and the automated execution of complex procedures~\cite{abdel2021self, dai2024autonomous, dragone2017autonomous}. 
Deploying embodied intelligence~\cite{billard2019trends,gong2023arnold} in such settings requires advanced capabilities, including long-horizon planning, fine-grained manipulation, and strict safety awareness.
However, collecting large-scale real-world data for training embodied agents in laboratories is often prohibitively expensive and risky, particularly for tasks involving delicate operations or high-cost failures. Thus, simulation and sim-to-real learning~\cite{zhao2020sim} serve as alternatives by offering scalable, safe, and controllable platforms for interactions and trial-and-error. 

Despite the impressive advancements in high-fidelity physics simulators~\cite{todorov2012mujoco,xiang2020sapien,hu2019difftaichi}, conventional simulation pipelines still rely heavily on teleoperation or manual specification~\cite{wang2024dexcap,fu2024mobile,gu2023maniskill2}. While such handcrafted designs are effective for controlled validation, they require substantial human effort and scale poorly in terms of task diversity and environment coverage. To alleviate this limitation, generative simulation~\cite{wang2023gensim,wang2023robogen,katara2024gen2sim,hua2024gensim2,gao2025genmanip} has recently emerged to automatically synthesize key components of simulation environments. 
However, existing approaches mainly target everyday manipulation and loosely structured interactions~\cite{li2023behavior,shridhar2020alfred}, limiting their applicability to scientific laboratory domains with highly structured, long-horizon tasks.
%
While recent works have introduced dedicated laboratory environments~\cite{li2025labutopia, lan2025autobio}, they primarily focus on isolated operation segments rather than protocol-aligned workflows.
Therefore, automatically synthesizing large-scale simulation data that faithfully follows experimental protocols for high-precision laboratory workflows remains an open question.

\begin{figure*}[t]
\begin{center}
\includegraphics[width=\textwidth]{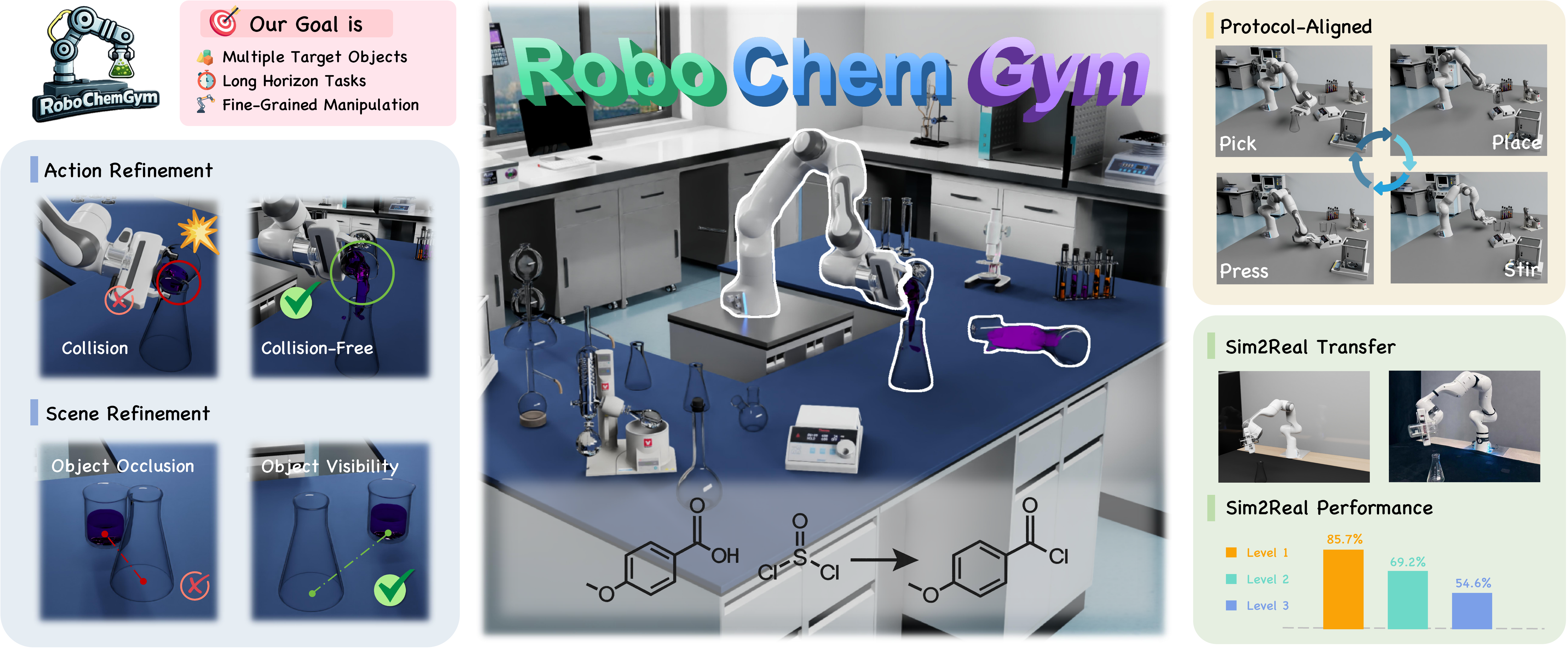}
\caption{%
    We illustrate a protocol-driven chemical manipulation task based on a real laboratory reaction.
    \textbf{Our Goal} is to support fine-grained, long-horizon manipulation involving multiple objects.
    \textbf{Action Refinement} corrects execution failures, while \textbf{Scene Refinement} resolves layout issues.
   \textbf{Protocol-Aligned}
    refers to manipulation trajectories generated by our framework for the given experimental protocol.
    Together, these components enable robust \textbf{Sim2Real} transfer.
    }
\label{fig:overview}
\end{center}
\end{figure*}

To address these challenges, we propose RoboChemGym, as illustrated in Figure~\ref{fig:overview}, an automated embodied manipulation framework tailored to real-world chemical experiment scenarios. It extends the principles of generative simulation to protocol-level task settings. RoboChemGym takes natural language chemical experiment protocols as input and automatically constructs simulation scenes and executable programs aligned with the experimental workflow. Given the strong coupling between actions and scenes in long-horizon laboratory workflows, independently optimizing either often results in infeasible executions or protocol violations. For example, a robotic arm cannot pour liquid into a beaker outside its reachable workspace while keeping a compliant posture. To overcome this, our framework adopts a memory-driven self-improving mechanism to iteratively refine action and scene, generating expert demonstration trajectories. Based on these expert trajectories, we further perform lab-centric data scaling by introducing variations in visual appearance, spatial configuration, and operational conditions within the scope of chemical experiment setups. This expands the simulated data distribution while maintaining chemical compliance.


Since chemical experiments consist of multiple interdependent steps with strict execution order, single-level evaluation metrics are insufficient to fully characterize an agent’s proficiency over complete experimental workflows. To reasonably evaluate robotic performance in the long-horizon, high-precision tasks, we design a hierarchical benchmark that derives tasks from real chemical experiment protocols. The benchmark is organized by increasing complexity, ranging from short sequences of $3$ atomic actions to full-scale experimental workflows comprising $11$ atomic actions. By evaluating success at both the atomic and procedural levels, it enables detailed analysis of error accumulation in long-horizon executions and facilitates standardized comparisons across methods.

We summarize our main contributions as follows: (1) We propose an automated framework for experiment generation and execution that faithfully follows real-world protocols, enabling scalable data synthesis strictly aligned with chemical procedures. (2) We present a self-improving task synthesis mechanism to reduce generation failures through joint action-scene optimization, ensuring robust generation performance in long-horizon tasks with dense object interactions. (3) We establish a hierarchical benchmark for embodied manipulation in chemical experiments, evaluating agent performance from atomic-level operations to complex, long-horizon workflows.

\section{Related Work}
\paragraph{Simulation-Based Robot Skill Learning.} Learning robot skills in simulation has become a foundational paradigm in embodied AI research~\cite{billard2019trends}, which accelerates data collection and mitigates real-world safety risks. The simulation platforms are generally categorized by the physical properties they simulate: rigid-body simulators like MuJoCo~\cite{todorov2012mujoco}, PyBullet~\cite{coumans2021}, and SAPIEN~\cite{xiang2020sapien}; deformable object simulators focusing on complex materials like cloth and rope~\cite{hu2019difftaichi, faure2012sofa}; and environments supporting multi-material coupling and fluid dynamics~\cite{wang2024thin, xian2023fluidlab, gu2023maniskill2}. These underlying engines provide the high-fidelity computational backbone necessary for robots to learn complex physical interactions. Leveraging these platforms, researchers have successfully trained a wide array of skills, evolving from basic ``pick-and-place" tasks with parallel grippers~\cite{tian2025interndata, liu2025fetchbot} to dexterous manipulation~\cite{wang2024dexcap, liang2025dexhanddiff, zhou2024learning}, dual arm operation~\cite{fu2024mobile, mu2024robotwin}, deformable object handling~\cite{scheikl2024movement, chi2024iterative} and legged locomotion~\cite{he2025attention, ha2025learning}.


\paragraph{Generative Simulation.} To overcome the inefficiency of manual environment design, generative simulation has emerged as an alternative paradigm. Recent works~\cite{yang2024holodeck, feng2023layoutgpt, fu2024anyhome} integrate Large Language Models (LLMs) with procedural generation to automatically map natural language descriptions to complex 3D scene layouts and asset generation. 
Frameworks like GenSim~\cite{wang2023gensim}, ReGen~\cite{nguyen2025regen} and RoboGen~\cite{wang2023robogen} utilize the code-generation capabilities of LLMs to achieve ``full-stack'' automation, covering task descriptions, layout optimization, and low-level control programs. 
This has vastly expanded the diversity of tasks available for robot learning. To further enhance robustness, researchers have introduced closed-loop feedback and self-evolving mechanisms: Eureka~\cite{ma2023eureka} automates the iteration of reward functions, while Voyager~\cite{wang2023voyager} enables continuous evolution of a skill library through code-error feedback. 
Existing generative frameworks are largely tailored for short-horizon, weakly constrained daily tasks, whereas specialized laboratory simulations~\cite{li2025labutopia, lan2025autobio} still depend on manual configuration. They struggle to handle the specialized requirements of chemical experiments, which demand adherence to rigorous protocols, coupling of multiple operational targets, and long-sequence reasoning.

\section{Protocol-Driven Data Generation}

\begin{figure*}[t]
\begin{center}
\centerline{\includegraphics[width=\linewidth]{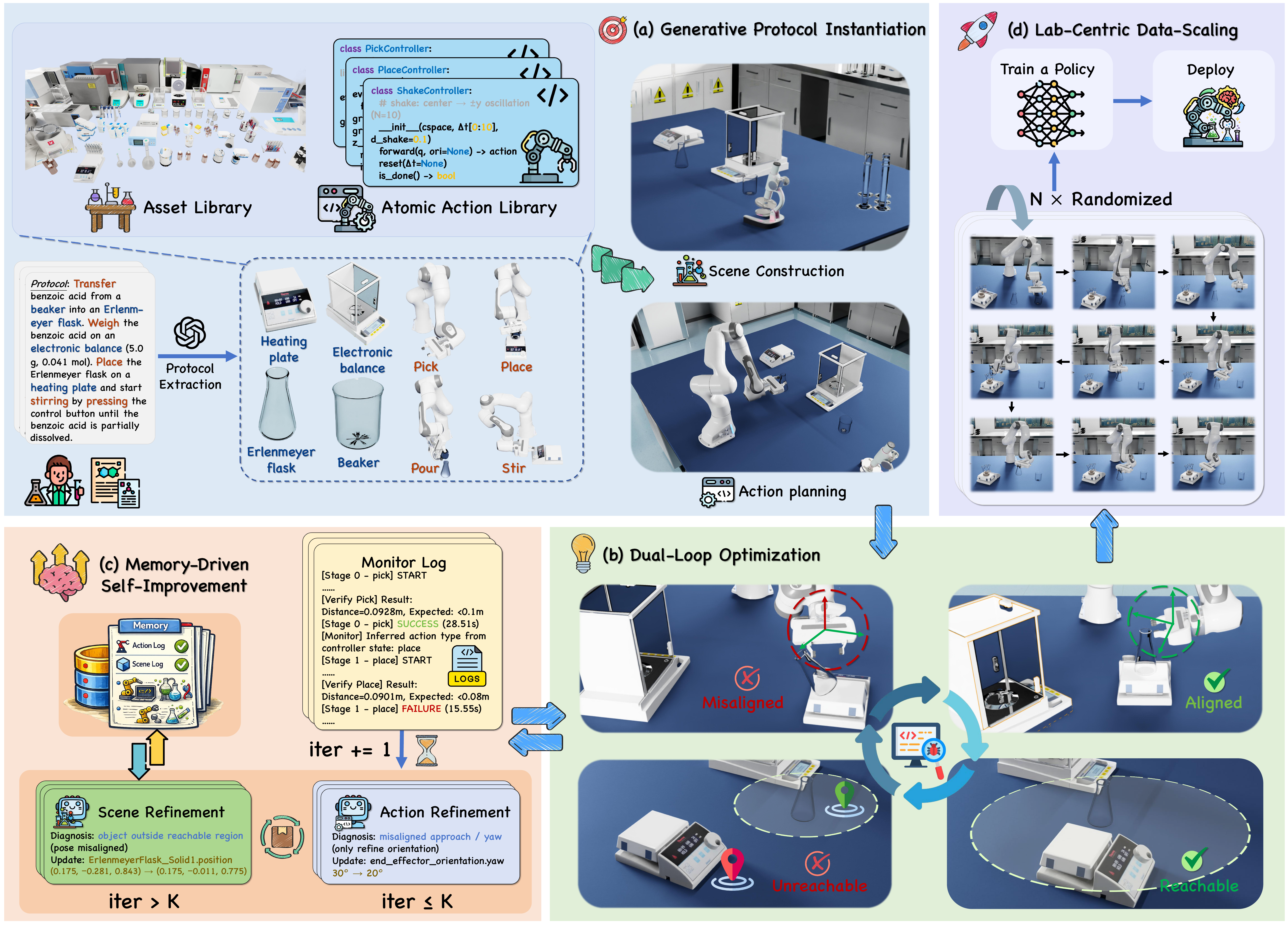}}
\caption{\textbf{RoboChemGym Pipeline.} \textbf{(a)} Given a textual protocol, the framework constructs an initial scene and executable action sequence using asset and atomic action libraries. \textbf{(b)} Dual-loop optimization refines actions and layouts under continuous monitoring, correcting execution errors and resolving global feasibility issues. \textbf{(c)} The memory module accumulates refinement histories to guide subsequent iterations and enable self-improvement. \textbf{(d)} Lab-centric data scaling applies randomization to generate diverse expert trajectories for policy training.}
\vspace{-15pt}
\label{fig:pipeline}
\end{center}
\end{figure*}

We propose a protocol-driven pipeline for automatically generating chemical experiment demonstrations, as shown in Figure~\ref{fig:pipeline}. Given a textual protocol, the framework produces diverse, executable task instances through three components:
1)\textit{Generative Protocol Instantiation} parses the protocol and constructs the initial scene and action sequence.
2)\textit{Self-Improving Task Synthesis} jointly refines scene layouts and action plans through memory-driven self-improvement to address complex experimental settings.
3)\textit{Lab-Centric Data Scaling} introduces protocol- and standard-aware visual and spatial variations, generating diverse demonstrations consistent with laboratory conventions and physical constraints for robust policy learning.


\subsection{Generative Protocol Instantiation}

Starting from a chemical protocol (\emph{\textit{e.g.}}, Place benzoic acid on a heating plate and initiate stirring \dots), we synthesize a high-quality tabletop experimental scene and executable trajectories that conform to the protocol. Specifically, we employ a large language model (Claude~3 Opus~\cite{claude3opus}; usage and cost details are provided in the Appendix) to extract scene descriptions and action sequences. Combined with an asset list and a predefined atomic action library, this structured information is passed to the \textit{Scene Construction Engine} and \textit{Action Planning Engine}.

\textbf{Scene Construction.}
We instantiate high-fidelity chemical assets from prior work~\cite{li2025labutopia} in realistic laboratory environments and formulate their placement as constrained optimization within the robot's reachable workspace. The workspace is estimated from sampled kinematic data, while layout optimization accounts for collisions, spatial distances, occlusion, and center preferences. The LLM further refines semantic relationships among assets to ensure consistency with real-world laboratory conventions. Detailed procedures are provided in the Appendix.

\textbf{Action Planning.}
Given the extracted action sequence, the Action Planning Engine retrieves the corresponding atomic actions and composes them into a complete executable program. The atomic action library, inherited from prior work~\cite{li2025labutopia}, contains eight primitives: \texttt{Pick, Place, Pour, Press, Open, Close, Shake}, and \texttt{Stir}. Each primitive is implemented as a parameterized interface that encapsulates an expert trajectory and exposes adjustable parameters for fine-grained control. In subsequent stages, we freeze non-parameterized code and optimize only these parameters, enabling the synthesis of diverse, high-fidelity expert demonstrations.


\subsection{Self-Improving Task Synthesis}

Although the initial generation provides a starting point, achieving collision-free and precise manipulation trajectories for multiple objects in a single pass proves computationally intractable for complex setups. Therefore, we propose a self-improving task synthesis mechanism to iteratively refine both task execution and scene configuration through successive optimization cycles. We formulate this approach as a closed-loop system comprising three core components: execution error monitoring, dual-loop optimization, and memory-driven self-improvement. Specifically, the error monitoring module continuously identifies execution failures and feeds this information to the optimization module, which iteratively refines either the scene or the manipulation trajectory in a dual-loop fashion by leveraging error signals and historical patterns stored in memory. The optimized configurations are committed back to memory, fostering a continuous cycle of adaptive improvement.




\textbf{Execution Error Monitoring.}
Unlike prior work~\cite{li2025labutopia} that only performs post hoc success checks for isolated atomic actions, we implement a continuous run-time error monitoring mechanism. This module segments the execution workflow into atomic stages based on controller states, while concurrently logging frame-level trajectories of the robot end-effector and all task-relevant objects. 
Within each stage, the monitor performs context-aware verification to generate diagnostic signals explicitly tailored for downstream correction. For instance, during a \texttt{pick} operation, it evaluates the geometric consistency between the gripper pose and the target grasp site; if the deviation exceeds a predefined threshold, a failure event is triggered. The resulting failure record provides comprehensive details, including the specific atomic action, the associated target object, quantitative metrics (\textit{e.g}., end-effector-to-target distance), and the termination context. By characterizing both the failure cause and its execution context, these structured records convert raw runtime errors into actionable semantic feedback, which is then directly consumed by the optimization module for targeted correction. It will also be stored in the memory module to support continual refinement of future task executions.

\textbf{Dual-Loop Optimization.}
We introduce a dual-loop optimization mechanism that jointly optimizes action strategies and scene configurations for multi-objective, high-precision robotic chemistry tasks. It consists of an inner action loop and an outer scene loop. The LLM diagnoses execution failures, generates corrective actions in the inner loop, and adjusts spatial layouts in the outer loop.
Upon receiving structured failure information from the error monitor, the system activates the inner loop, where the LLM iteratively refines low-level action parameters using current failure signals and historical corrective patterns retrieved from action memory.
If failures persist after several rounds of refinement, the system enters the outer loop. The LLM then adjusts the scene configuration based on observed failure patterns and relevant scene-memory experiences. For example, repeated \texttt{Pick} failures with consistent gripper offsets may indicate that the target is outside the robot's reachable workspace and should be repositioned to a feasible pose.
The two loops operate collaboratively to reduce execution failures until the predefined success criteria are met.



\textbf{Memory-Driven Self-Improvement.}
%
Given the sequential dependencies in chemical protocols, we use insights from each iteration to guide future decisions. This accumulated experience enables the system to learn from failures and reduce error propagation. To support continuous learning, we introduce a structured memory system that archives the LLM's analysis and optimization results. The Action Memory stores textual diagnoses of failed actions, including relative pose deviations between the gripper and object and subsequent parameter adjustments. The Scene Memory records spatial reconfigurations caused by environmental constraints or object-specific properties (\textit{e.g.}, clearance requirements for container opening). By combining retrieved experience with real-time execution feedback, the LLM dynamically adapts its decisions, establishing a closed-loop self-improvement mechanism that progressively enhances task performance.

\subsection{Lab-Centric Data Scaling}

To improve policy robustness to real-world stochasticity and reduce the sim-to-real gap, we employ a multi-dimensional data scaling strategy that incorporates the operational rules and constraints of chemical experiments. Specifically, we perturb the data distribution along four axes: 1) \textit{spatial stochasticity} of manipulated objects, 2) \textit{background distractions}, 3) \textit{material heterogeneity} of laboratory workbenches, and 4) \textit{illumination variability}. Further details are provided in the Appendix.

\textbf{Spatial Stochasticity.}
Chemical experiments often require sequential manipulation of multiple objects, demanding robust long-horizon policies. We therefore apply controlled position randomization to each target asset, diversifying manipulation trajectories and visual observations while preserving safety and task feasibility.

\textbf{Background Distractions.}
To better reflect real laboratory environments and improve visual robustness, we add distractor assets to the workbench. Although unrelated to the current task, these assets remain semantically consistent with the experimental context and are inserted using collision-aware placement to ensure physical plausibility.

\textbf{Material Heterogeneity.}
To represent the diverse surfaces found in laboratories, including workbenches and fume hoods, we introduce a high-fidelity texture library and randomize workbench textures. The library combines official Isaac Sim textures with over a dozen curated laboratory bench textures, increasing visual diversity and realism.

\textbf{Illumination Variability.}
Laboratory lighting varies in color temperature, source type, intensity, and spatial arrangement, affecting the appearance and reflectance of experimental instruments. We randomize lighting color, type, intensity, and position within physically plausible ranges, exposing policies to realistic illumination variations and improving robustness to real-world lighting changes.

\section{Protocol-Aligned Hierarchical Benchmark}
%
Leveraging the proposed automated data generation pipeline, we construct a diverse suite of $20$ tasks from realistic chemical protocols, including $9$ short-horizon, $8$ long-horizon, and $3$ ultra-long-horizon tasks. The tasks vary substantially in manipulation targets and action-sequence lengths.
To systematically evaluate embodied agents, we establish a protocol-aligned hierarchical benchmark covering reasoning, planning, and control. More than half of the tasks contain over five atomic actions, providing a standardized benchmark for long-horizon robotic manipulation in chemical environments. Detailed task statistics and protocol definitions are provided in the Appendix.






\subsection{Hierarchical Task Setup}

We categorize manipulation tasks into three levels based on action horizon and manipulation object complexity. Level 1 consists of short-horizon, single-target tasks involving $3$ atomic actions, focusing on precise execution and smooth coordination across consecutive actions. Level 2 extends to long-horizon tasks with multiple targets, comprising $7$ atomic actions and two to three manipulation objects. These tasks require the agent to exhibit high-level planning and reasoning capabilities in order to maintain execution accuracy and efficiency during target switching. Level 3 further scales to ultra-long-horizon tasks with $11$ atomic actions and more than three manipulation objects, placing greater demands on the agent’s memory and long-term task planning to ensure consistency and coherence over extended action sequences. This hierarchical task design enables a systematic evaluation of embodied agents in chemical experimental settings, particularly for long-horizon, fine-grained manipulation involving multiple interacting objects.

\subsection{Multi-Granularity Long-Horizon Evaluation}

Considering the inherent path dependency of chemical experiments (\textit{$i.e.$}, the outcome of preceding actions directly influences the success of subsequent actions), we introduce a multi-granularity evaluation protocol that assesses both the atomic action level and the full task level. 
At the atomic action level, we define explicit goal states and corresponding tolerance thresholds for each atomic operation. 
%
%
At the full task level, a task is considered successful only if all atomic actions in the sequence are individually successful.


\section{Experiments}
We conducted a series of experiments to evaluate the effectiveness of our method across three key dimensions: (i) the automated generation of expert trajectories for chemical experiments; (ii) the evaluation of various manipulation policies with lab-centric data scaling; and (iii) evaluation on sim-to-real performance. In our experiments, the clean configuration refers to data collected without lab-centric data scaling, while the random configuration refers to data augmented with variations from the lab-centric data scaling process.
\definecolor{amviolet}{HTML}{E6E6FF}
\newcommand{\tagAM}{\cellcolor{amviolet}$-$AM}
\newcommand{\tagSM}{\cellcolor{amviolet}$-$SM}

\newcommand{\dec}[1]{%
  \hspace{2pt}%
  \colorbox{yellow!25}{\raisebox{0pt}[0.9\height][0.8\depth]{\scriptsize$\downarrow$#1}}%
}

\begin{table}[t]

\begin{center}
\begin{small}

\setlength{\tabcolsep}{6pt}
\renewcommand{\arraystretch}{1.12}

\begin{tabular}{ccccc}
\toprule
Setting & Config & Level 1 & Level 2 & Level 3 \\
\midrule
\multirow{3}{*}{Clean}   & Ours        & \textbf{89.2} & \textbf{73.4} & \textbf{52.3} \\
                         & $-$AM        & \underline{75.8}\dec{13.4} & 52.7\dec{20.7} & 27.4\dec{24.9} \\
                         & $-$SM        & 73.6\dec{15.6} & \underline{56.4}\dec{17.0} & \underline{32.3}\dec{20.0} \\
\midrule
\multirow{3}{*}{Random}  & Ours        & \textbf{85.6} & \textbf{70.3} & \textbf{47.5} \\
                         & $-$AM        & \underline{69.7}\dec{15.9} & 48.9\dec{21.4} & 21.2\dec{26.3} \\
                         & $-$SM        & 64.1\dec{21.5} & \underline{51.8}\dec{18.5} & \underline{25.4}\dec{22.1} \\
\bottomrule
\end{tabular}
\caption{Success rates (\%) of the generative pipeline across configurations and task difficulty levels. \textit{Clean} excludes lab-centric data scaling, while \textit{Random} includes it. \textbf{-AM} and \textbf{-SM} denote removing action- and scene-level memory, respectively. The $\Delta$ values indicate the corresponding decreases in success rate.
}
\label{tab:pipeline_success}
\end{small}
\end{center}
\vskip -0.2in
\end{table}



\subsection{Evaluation of Automated Task Generation}

We evaluate the success rate of the proposed generative pipeline on $20$ chemical experiment protocols. The pipeline automatically generates experimental scenes and action sequences, with trajectories collected through repeated executions and randomization strategies. Success rate is defined as the ratio of valid trajectories to total attempts, with results summarized in Table~\ref{tab:pipeline_success}.

With up to 30 iterations, the proposed pipeline demonstrates consistent stability. While success rates naturally scale with task complexity due to inevitable error accumulation in long-horizon workflows, our method maintains competitive performance in the clean setting, achieving success rates of 89.2\%, 73.4\%, and 52.3\% across the three difficulty levels.
Under randomized conditions, we observe only a marginal performance drop, averaging 3.83\% (with success rates of 85.6\%, 70.3\%, and 47.5\%), which highlights the robustness and strong generalization capabilities of our approach under both visual and dynamic uncertainties. We further assess the generated trajectories through real-robot replay. Without reset actions, full-sequence success rates are 85.7\%, 69.2\%, and 54.6\% for Levels~1--3, respectively. With reset actions inserted after each pick-and-place pair, these rates increase to 87.3\%, 79.7\%, and 70.5\%. Further details are provided in Appendix~\ref{app:Sim2real}
\begin{table*}[t]

\centering
\begin{tabular}{lcccccccccccc}
\toprule
& \multicolumn{4}{c}{\textbf{Level-1 (3-step)}} & \multicolumn{8}{c}{\textbf{Level-2 (7-step)}} \\
\cmidrule(lr){2-5}\cmidrule(lr){6-13}
\textbf{Method} & \textbf{A1} & \textbf{A2} & \textbf{A3} & \textbf{Full Seq} &
\textbf{A1} & \textbf{A2} & \textbf{A3} & \textbf{A4} & \textbf{A5} & \textbf{A6} & \textbf{A7} & \textbf{Full Seq} \\
\midrule
$\pi_0$ & 14.3 & 4.7 & 1.3 & 0.8 & 11.5 & 3.1 & 0.8 & 0.2 & 0 & 0 & 0 & 0 \\
$\pi_0{+}\text{clean}$ & {90.2} & {82.5} & {75.6} &
{55.4 {\tiny\textcolor{ForestGreen}{↑ 54.6}}} &
84.5 & 72.1 & 60.4 & 50.2 & 42.1 & 30.5 & 22.4 &
{18.2 {\tiny\textcolor{ForestGreen}{↑ 18.2}}} \\
$\pi_0{+}\text{random}$ & {94.1} & {88.6} & {82.3} &
{68.5 {\tiny\textcolor{ForestGreen}{↑ 67.7}}} &
88.5 & 78.4 & 68.2 & 58.4 & 48.5 & 40.2 & 32.5 &
{28.4 {\tiny\textcolor{ForestGreen}{↑ 28.4}}} \\
ACT$+$clean & {72.0} & {62.5} & {50.4} & {22.5} &
65.5 & 52.3 & 40.1 & 32.5 & 24.8 & 15.2 & 8.5 & 5.2 \\
ACT$+$random & {80.5} & {70.2} & {60.1} & {35.2} &
72.1 & 58.4 & 45.2 & 38.6 & 30.1 & 22.4 & 15.8 & 12.4 \\
DP$+$clean & {65.2} & {55.1} & {42.3} & {15.4} &
55.4 & 40.2 & 30.5 & 22.1 & 15.6 & 10.2 & 5.4 & 3.8 \\
DP$+$random & {75.1} & {65.3} & {52.8} & {25.6} &
66.2 & 50.1 & 38.4 & 30.2 & 22.5 & 15.3 & 10.1 & 8.5 \\
\bottomrule
\end{tabular}
\caption{Step-wise and full-sequence success rates (\%) in long-horizon chemical manipulation tasks. 
Level-1 contains 3 actions and Level-2 contains 7 actions. 
{↑} is reported for $\pi_0{+}\text{clean}$ and $\pi_0{+}\text{random}$, denoting absolute improvement over $\pi_0$ on \textit{Full Seq} metric.}
\label{tab:long_horizon_chemical_tasks}
\end{table*}


\begin{table*}[t]

\centering
\begin{tabular}{lcccccccccccc}
\toprule
& \multicolumn{12}{c}{\textbf{Level-3 (11-step)}} \\
\cmidrule(lr){2-13}
\textbf{Method} & \textbf{A1} & \textbf{A2} & \textbf{A3} & \textbf{A4} & \textbf{A5} & \textbf{A6} & \textbf{A7} & \textbf{A8} & \textbf{A9} & \textbf{A10} & \textbf{A11} & \textbf{Full Seq} \\
\midrule
$\pi_0$ & 9.5 & 2.1 & 0.5 & 0 & 0 & 0 & 0 & 0 & 0 & 0 & 0 & 0 \\
$\pi_0{+}\text{clean}$ & 84.4 & 72.0 & 60.3 & 50.1 & 42.0 & 30.4 & 22.3 & 16.1 & 11.6 & 8.3 & 6.0 &
0.5 {\scriptsize\textcolor{ForestGreen}{↑ 0.5}} \\
$\pi_0{+}\text{random}$ & 88.4 & 78.3 & 68.1 & 58.3 & 48.4 & 40.1 & 32.4 & 23.3 & 16.8 & 12.1 & 8.7 &
1.7 {\scriptsize\textcolor{ForestGreen}{↑ 1.7}} \\
\bottomrule
\end{tabular}
\caption{Step-wise and full-sequence success rates (\%) in ultra-long horizon chemical manipulation tasks. Level-3 contains 11 actions. \textcolor{ForestGreen}{↑} is reported for $\pi_0{+}\text{clean}$ and $\pi_0{+}\text{random}$, denoting absolute improvement over $\pi_0$ on \textit{Full Seq} metric.}
\label{tab:ultra_long_horizon_chemical_tasks}
\end{table*}

\textbf{Effects of Memory Mechanism.} We perform a comparative analysis of our generation pipeline by removing the action-level feedback memory (\textit{w/o} AM) and removing the scene-level feedback memory (\textit{w/o} SM), respectively. The results are summarized in Table~\ref{tab:pipeline_success}. 
The results indicate that removing either memory module leads to performance degradation. Notably, the impact of action memory intensifies significantly with task difficulty: for Level-1 short-horizon tasks, removing AM reduces the success rate from $89.2$\% to $75.8$\%, whereas for Level-3 ultra-long-horizon tasks, it plummets from $52.3$\% to $27.4$\% under the clean setting. This highlights the critical role of historical action-level information in suppressing error accumulation during long-sequence execution.

\begin{figure}[t]
    \centering
    \includegraphics[width=\columnwidth]{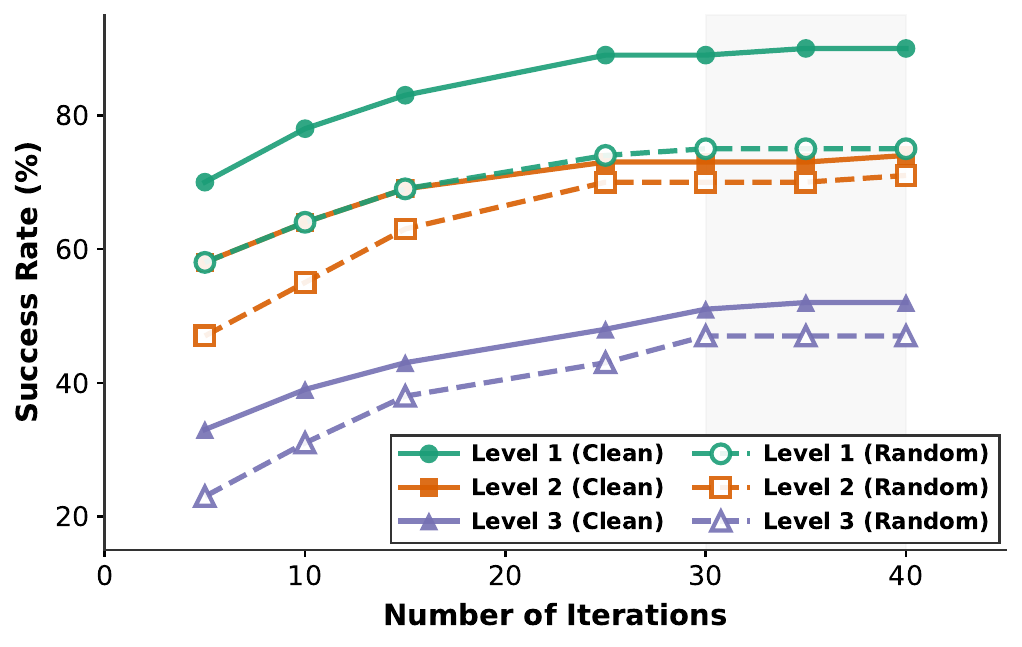}
    \caption{Success rates of Level 1, Level 2, and Level 3 under clean and random settings as a function of the number of iterations.}
    \label{fig:pipeline_success_rate}
    \vspace{-10pt}
\end{figure}
\textbf{Effects of the Iterations.} We analyze the effect of iteration depth on task generation success rates, with results summarized in Figure~\ref{fig:pipeline_success_rate}. The results show that success rates across all difficulty levels increase with the number of iterations, but exhibit diminishing returns beyond a certain iteration depth. For Level-3 tasks, the success rate improves from $33$\% in early iterations to around $48$\% under the Clean setting, and eventually stabilizes at $52$\%; a similar plateau at approximately $47$\% is observed under the Random setting. These results indicate that the dual-feedback memory mechanism primarily corrects critical deviations during early iterations; once the system approaches a feasible solution, subsequent iterations yield only limited improvements, with final performance constrained by task complexity and environmental uncertainty. We analyze generation failures and find that in simple scenarios (Level 1–2, fewer than $4$ objects), errors mainly arise from action failures caused by parameter deviations in atomic actions, which can be corrected through iterative refinement. In contrast, in complex scenarios (Level 3, $4$–$5$ objects), failures primarily stem from the complexity of multi-object layouts, such as arranging multiple objects within the limited workspace while ensuring sequential interactions without occlusion or collision. These compounded constraints make the task challenging even for human experts.



\vspace{-2pt}
\subsection{Evaluation of Various Manipulation Policies}


Our goal is to evaluate whether the expert trajectories generated by the proposed data collection method can effectively improve the robustness of strategies in multi-step, fine-grained chemical manipulation tasks. To this end, we collect $200$ expert trajectories for each task under the clean and random setups, respectively. 
We select three representative models: ACT~\cite{fu2024mobilealohalearningbimanual},
Diffusion Policy (DP)~\cite{chi2025diffusion}, and $\pi_0$~\cite{black2024pi_0}.
The models are evaluated in randomized environments that are not seen during training, with variations in object placement, background configuration, and illumination conditions. Training and inference details for policies are provided in Appendix.
\begin{table*}[t]
\centering
\small
\setlength{\tabcolsep}{4pt}

\begin{tabular}{
>{\centering\arraybackslash}m{2.7cm}
>{\centering\arraybackslash}m{2.5cm}
>{\centering\arraybackslash}m{1.0cm}
>{\centering\arraybackslash}m{1.5cm}
>{\centering\arraybackslash}m{1.6cm}
>{\centering\arraybackslash}m{2.2cm}
}

\toprule

\textbf{Task} &
\textbf{Position} &
\textbf{Clutter} &
\textbf{Clean Real} &
\textbf{RoboChemGym} &
\textbf{Combined}
\\

\midrule

\multirow{4}{*}{\centering Solid Weighing}
&
\multirow{2}{*}{\centering Seen-Position}
& No
& 19.8
& 20.5
& 41.8
\\

&
& Yes
& 14.9
& 17.2
& 38.5
\\

\cmidrule(lr){2-6}

&
\multirow{2}{*}{\centering Unseen-Position}
& No
& 13.2
& 16.8
& 36.9
\\

&
& Yes
& 9.1
& 13.5
& 32.7
\\

\midrule

\multirow{4}{*}{\centering Pyridine Addition}
&
\multirow{2}{*}{\centering Seen-Position}
& No
& 23.5
& 29.3
& 36.4
\\

&
& Yes
& 18.1
& 22.4
& 33.6
\\

\cmidrule(lr){2-6}

&
\multirow{2}{*}{\centering Unseen-Position}
& No
& 15.6
& 21.9
& 31.5
\\

&
& Yes
& 10.8
& 18.2
& 28.9
\\

\midrule

\multirow{4}{*}{\centering pH Measurement}
&
\multirow{2}{*}{\centering Seen-Position}
& No
& 15.2
& 24.7
& 47.6
\\

&
& Yes
& 13.5
& 20.3
& 44.8
\\

\cmidrule(lr){2-6}

&
\multirow{2}{*}{\centering Unseen-Position}
& No
& 11.8
& 17.5
& 42.1
\\

&
& Yes
& 8.7
& 14.9
& 38.6
\\

\midrule

\multirow{4}{*}{\centering \textbf{Average}}
&
\multirow{2}{*}{\centering Seen-Position}
& No
& 19.5
& 24.8
& \textbf{41.9 (+22.4)}
\\

&
& Yes
& 15.5
& 20.0
& \textbf{39.0 (+23.5)}
\\

\cmidrule(lr){2-6}

&
\multirow{2}{*}{\centering Unseen-Position}
& No
& 13.5
& 18.7
& \textbf{36.8 (+23.3)}
\\

&
& Yes
& 9.5
& 15.5
& \textbf{33.4 (+23.9)}
\\

\bottomrule

\end{tabular}

\caption{
Real-world success rates (\%) under different task configurations.
Seen-Position and Unseen-Position indicate whether object locations are observed during training.
Clutter indicates the presence of distractor objects.
Numbers in parentheses denote absolute improvements.
}
\label{tab:robochemgym_level1_final_percent}
\vspace{-7pt}

\end{table*}

We report two types of metrics: the success rate of the entire action sequence and the independent success rate of individual actions (\textit{i.e}., the probability of success for each action given that all previous actions are successful). This evaluation method provides a more comprehensive picture of the strategy's performance in different operational stages of long-horizon tasks. For $\pi_0$, we additionally evaluate its performance in ultra-long sequence tasks at Level-3.

\textbf{Performance Comparison Across Policies and Task Horizons.} As shown in Tables~\ref{tab:long_horizon_chemical_tasks} and~\ref{tab:ultra_long_horizon_chemical_tasks}, policy performance differs substantially on long-horizon chemical manipulation tasks. Under the same data settings, ACT and DP achieve full-sequence success rates of $22.5$\% and $15.4$\% on Level-1 tasks, but their performance degrades rapidly with increasing horizon, dropping to $5.2$\% and $3.8$\% on Level-2 tasks and to near $0$ on ultra-long Level-3 tasks. In contrast, $\pi_0$-based policies are more stable on longer horizons: after training on expert trajectories collected by our method, $\pi_0$ achieves $18.2$\% (\text{Clean}) and $28.4$\% (\text{Random}) full-sequence success on Level-2 tasks, and $\pi_0{+}\text{random}$ maintains a $1.7$\% success rate on Level-3 tasks. Compared with ACT and DP, $\pi_0$ consistently maintains higher step-wise success rates across all action stages in both Level-2 and Level-3 tasks. Notably, although long-horizon tasks remain limited by model capacity and thus exhibit low overall success rates (e.g., $\pi_0$ achieves 0\% zero-shot on Level-3, increasing to 0.5–1.7\% after training), training with RoboChemGym data still significantly improves the success rates of most individual actions within the sequence, demonstrating its effectiveness for ultra-long task training.


\textbf{Effect of Lab-Centric Data Scaling.} Results in Table~\ref{tab:long_horizon_chemical_tasks} and~\ref{tab:ultra_long_horizon_chemical_tasks} highlight the efficacy of our lab-centric data scaling strategy, especially for long-horizon tasks. Across all difficulty levels, models trained with randomized data consistently outperform their counterparts, with the performance gap widening as task complexity increases. For instance, $\pi_{0}$ exhibits substantial gains on Level-1 ($55.4$\% $\rightarrow$ $68.5$\%) and Level-2 ($18.2$\% $\rightarrow$ $28.4$\%) tasks, while also showing relative improvements on the challenging Level-3 tasks. Moreover, $\pi_0{+}\text{random}$ maintains higher step-wise success rates than $\pi_0{+}\text{clean}$ across all action stages in Level-2 and Level-3 tasks, with larger gains in later stages. These results demonstrate that lab-centric data scaling effectively reduces error buildup and enhances robustness in long-horizon chemical manipulation.

\subsection{Evaluation on Sim-to-Real Performance}
To evaluate RoboChemGym's effectiveness in improving policy robustness, we conduct experiments on three Level-1 tasks (following RoboTwin~\cite{mu2024robotwin}). All experiments use $\pi_0$ as the policy backbone. We compare three training settings: (i) 10 real demonstrations on a clean tabletop; (ii) a zero-shot setting with 1,000 synthetic trajectories; and (iii) augmenting the 10 real demonstrations with 1,000 synthetic trajectories through domain randomization, introducing variations like positional generalization and cluttered scenes.

Evaluation is conducted under four configurations: (i) clean tabletop with in-domain locations, (ii) clean tabletop with seen positions, (iii) cluttered tabletop with in-domain locations, and (iv) cluttered tabletop with seen positions. This setup tests whether RoboChemGym enables robust policy generalization without additional real-world data from seen positions configurations.

The experimental results, as shown in Table~\ref{tab:robochemgym_level1_final_percent}, demonstrate that policies augmented with RoboChemGym achieve significant improvements in robustness on real-world tasks. When $1,000$ domain-randomized synthetic trajectories are combined with only $10$ real-world demonstrations, the average success rate across all evaluation configurations increases by $23.3$\%, with configuration-specific gains of $22.4$\%, $23.5$\%, $23.3$\%, and $23.9$\%, respectively. In the zero-shot setting trained solely on synthetic data, moderate improvements are still observed under test scenarios, reaching $5.3$\%, $4.5$\%, $5.2$\%, and $6.0$\%, respectively. Notably, gains are more pronounced in cluttered, visually complex scenes, indicating that RoboChemGym enhances robustness in challenging real-world conditions.

These improvements stem from two factors: (i) RoboChemGym provides high visual and physical fidelity, enabling policies to transfer directly from simulation to the real world; and (ii) domain-randomized synthetic trajectories compensate for limited environmental variability in real demonstrations, allowing policies to adapt in advance to diverse real-world scenarios. Overall, even a small number of real demonstrations can bridge the sim-to-real gap, highlighting RoboChemGym’s advantage in generalization.


\vspace{-3pt}
\section{Conclusion \& Limitations}
We propose an automated embodied manipulation framework that extends generative simulation to chemical experiment tasks. By leveraging LLMs to convert textual protocols into high-fidelity, long-horizon expert demonstrations, our approach enables scalable data generation in complex laboratory workflows.
Although our system excels in chemical manipulation tasks, it has several limitations:
%
%
(1) While our current evaluation extensively utilizes simulation to ensure reproducibility and scalability across diverse tasks, extending real-world validation from trajectory replay to fully autonomous execution remains a valuable direction for future exploration.
(2) RoboChemGym currently supports only the Franka Panda; extending additional robot embodiments would enhance generalization in laboratory automation. 

\bibliography{aaai2027}
\newpage
\onecolumn

\section{Computational Cost Analysis}
\label{app:cost_analysis}

This section analyzes the computational overhead introduced by LLM API calls in our
protocol-driven generative simulation framework.
All statistics are obtained from empirical logs collected during actual optimization runs,
where each log entry corresponds to a single API call.

\subsection{Logging Setup and Task-Level Aggregation}

LLM calls are invoked exclusively for refinement purposes, including
(i) localized \emph{action refinement} conditioned on execution failures or monitor feedback, and
(ii) sparse \emph{scene refinement} triggered by reachability, collision, or occlusion violations.
Tasks are grouped into three difficulty levels according to protocol length and horizon:
\textbf{Level~1} (short horizon), \textbf{Level~2} (long horizon), and \textbf{Level~3} (ultra-long horizon).

Empirically, tasks at Level~1, Level~2, and Level~3 require approximately
$\sim$15, $\sim$20, and $\sim$30 LLM calls, respectively.
All per-task statistics reported below are obtained by aggregating empirical per-call measurements
according to these average call counts.

\subsection{Empirical Call Patterns}

Inspection of the collected logs reveals three dominant categories of API calls.

\paragraph{Short calls.}
A large fraction of calls fall into a short-call regime, with input lengths of approximately
$1337$ tokens and output lengths in the range of $1360$--$1530$ tokens.
These calls typically correspond to lightweight action-level fixes, such as end-effector pose adjustments,
parameter corrections, or short code fragments.

\paragraph{Medium calls.}
Another substantial portion of calls exhibit input lengths of approximately $2400$--$2600$ tokens,
with output lengths around $2200$--$2550$ tokens.
Such calls generally involve richer contextual updates, incorporating recent execution traces,
constraint violations, and partial scene states.

\paragraph{Heavy calls.}
A small number of calls involve significantly longer inputs (e.g., $18$k, $22$k, $50$k, or $69$k tokens),
while their outputs remain comparatively bounded ($\sim$3k--8k tokens).
These calls are associated with sparse global reasoning steps, such as scene reconfiguration
or protocol-level restructuring, and are invoked infrequently.

Across both short and medium calls, the input and output token counts are often of comparable magnitude.
This behavior is expected in our dual-loop refinement setting, where the model is tasked with transforming
structured execution logs, state descriptions, and constraints into correspondingly structured repair
suggestions, code snippets, or action sequences.

\subsection{Per-Task Token Footprint and Cost}

Table~\ref{tab:llm_cost_per_task} summarizes the estimated per-task LLM overhead for different
task levels.
We report both \emph{Low} and \emph{Typical} estimates, reflecting the two most common call regimes
observed in the empirical logs.

\begin{table}[H]
\centering
\small
\caption{Per-task LLM overhead for protocol-driven refinement.
Each task triggers multiple API calls (Level~1/2/3: $\sim$15/20/30 calls).
\emph{Low} and \emph{Typical} estimates are derived from the two most frequent call patterns observed
in empirical logs (one record corresponds to one API call).
TTFW is measured under non-streaming settings.}
\label{tab:llm_cost_per_task}
\setlength{\tabcolsep}{5pt}
\begin{tabular}{lccccc}
\toprule
\textbf{Task Level} &
\textbf{\# Calls} &
\textbf{Input Tokens / Task} &
\textbf{Output Tokens / Task} &
\textbf{Cost / Task (USD)} &
\textbf{TTFW / Call (s)} \\
\midrule
Level~1 &
$\sim$15 &
\begin{tabular}[c]{@{}c@{}}Low: $\sim$20k\\ Typical: $\sim$38k\end{tabular} &
\begin{tabular}[c]{@{}c@{}}Low: $\sim$21k\\ Typical: $\sim$36k\end{tabular} &
\begin{tabular}[c]{@{}c@{}}Low: $\sim$0.29\\ Typical: $\sim$0.50\end{tabular} &
\begin{tabular}[c]{@{}c@{}}$\sim$15--20\\ $\sim$25--40\end{tabular} \\
\midrule
Level~2 &
$\sim$20 &
\begin{tabular}[c]{@{}c@{}}Low: $\sim$27k\\ Typical: $\sim$50k\end{tabular} &
\begin{tabular}[c]{@{}c@{}}Low: $\sim$28k\\ Typical: $\sim$48k\end{tabular} &
\begin{tabular}[c]{@{}c@{}}Low: $\sim$0.38\\ Typical: $\sim$0.66\end{tabular} &
\begin{tabular}[c]{@{}c@{}}$\sim$15--20\\ $\sim$25--40\end{tabular} \\
\midrule
Level~3 &
$\sim$30 &
\begin{tabular}[c]{@{}c@{}}Low: $\sim$40k\\ Typical: $\sim$75k\end{tabular} &
\begin{tabular}[c]{@{}c@{}}Low: $\sim$42k\\ Typical: $\sim$72k\end{tabular} &
\begin{tabular}[c]{@{}c@{}}Low: $\sim$0.57\\ Typical: $\sim$0.99\end{tabular} &
\begin{tabular}[c]{@{}c@{}}$\sim$15--20\\ $\sim$25--40\end{tabular} \\
\bottomrule
\end{tabular}
\end{table}

\subsection{Latency Considerations}
\label{sec:cost_latency}

We report \emph{Time-to-First-Word} (TTFW) as the primary latency indicator, measured under non-streaming
settings.
Empirically, per-call TTFW typically falls in the $15$--$40$\,s range for short and medium calls,
with occasional larger values for heavy calls involving long-context prefill.
Task-level wall-clock latency is not reported as a simple sum over calls, since it also depends on
generation duration, environment rollouts, and server-side queueing variability.

\subsection{Discussion}

Despite involving multiple LLM calls per task, the overall computational overhead remains modest.
Even for ultra-long-horizon Level~3 tasks, the total monetary cost is on the order of one US dollar per task,
and the aggregate token footprint remains well below the context limits of modern large language models.
These results reflect a deliberate design choice: our framework relies on localized, structured refinements
rather than repeatedly invoking global, end-to-end reasoning over entire protocols.
Expensive long-context calls are triggered only sparsely, allowing computational cost to scale
gracefully with task difficulty and making our approach practical for large-scale simulation-based
data generation.

\section{Scene Construction Details}
\label{app:Scene Construction}

We construct initial laboratory scenes using two key resources: a library of over $140$ high-fidelity chemical assets and a collection of more than $10$ laboratory environments that replicate real-world layouts, materials, and lighting conditions. Based on the extracted scene specifications, the Scene Construction Engine retrieves all required assets and instantiates them within a laboratory environment to match the target experimental setup.

Asset placement is formulated as a constrained optimization problem. The robot’s reachable workspace is estimated from sampled kinematic data and modeled as an elliptical planar region, which defines the feasible placement area for all assets. Within this region, the layout must satisfy several constraints, including collision avoidance, minimum inter-asset distance requirements, clear mutual visibility, and alignment with operational center preferences.

These constraints are integrated into a weighted objective function to efficiently search for feasible and near-optimal arrangements, even in scenes containing a large number of assets. Formally, the optimization problem is defined as:
\begin{align}
\label{eq:scene_layout_appendix}
\min_{x_i} \Big( 
& w_1 \sum_{i \neq j} |x_i  x_j| 
+ w_2 \sum_i \mathbf{d}(A_i) \notag \\
& + w_3 \sum_{i \neq j} \big(\mathbb{V}(A_i, A_j)\big) 
\Big)
\end{align}
\begin{equation*}
\text{s.t.} \quad d(A_i, A_j) \geq \delta, \quad \forall i \neq j .
\end{equation*}

Here, $x_i$ denotes the 3D position of asset $A_i$ in the scene. $\mathbb{V}(A_i, A_j)$ measures pairwise visibility between assets, where $\mathbb{V}(A_i, A_j)=1$ indicates full mutual visibility and $\mathbb{V}(A_i, A_j)=0$ indicates complete occlusion. $\mathbf{d}(A_i)$ represents the distance from asset $A_i$ to the operational center, and $d(A_i, A_j)$ enforces a minimum safety margin $\delta$. The weights $w_1$, $w_2$, and $w_3$ balance spatial separation, center alignment, and visibility considerations, respectively.

Furthermore, \( w_1 \), \( w_2 \), and \( w_3 \) are set to $0.75$, $0.6$, and $0.35$, respectively, based on extensive experimental validation. These values prove effective in achieving the desired outcomes. Additionally, considering both the operational range preferences of the robotic arm and extensive experimental data, we set the minimum distance \( d(A_i, A_j) \) between objects to $15$ cm, ensuring a reasonable spatial separation.

\section{Data-Scaling and Visual Variations}
\label{app:Domain Randomization}

\begin{figure}[H]
    \centering
    \includegraphics[width=0.6\linewidth]{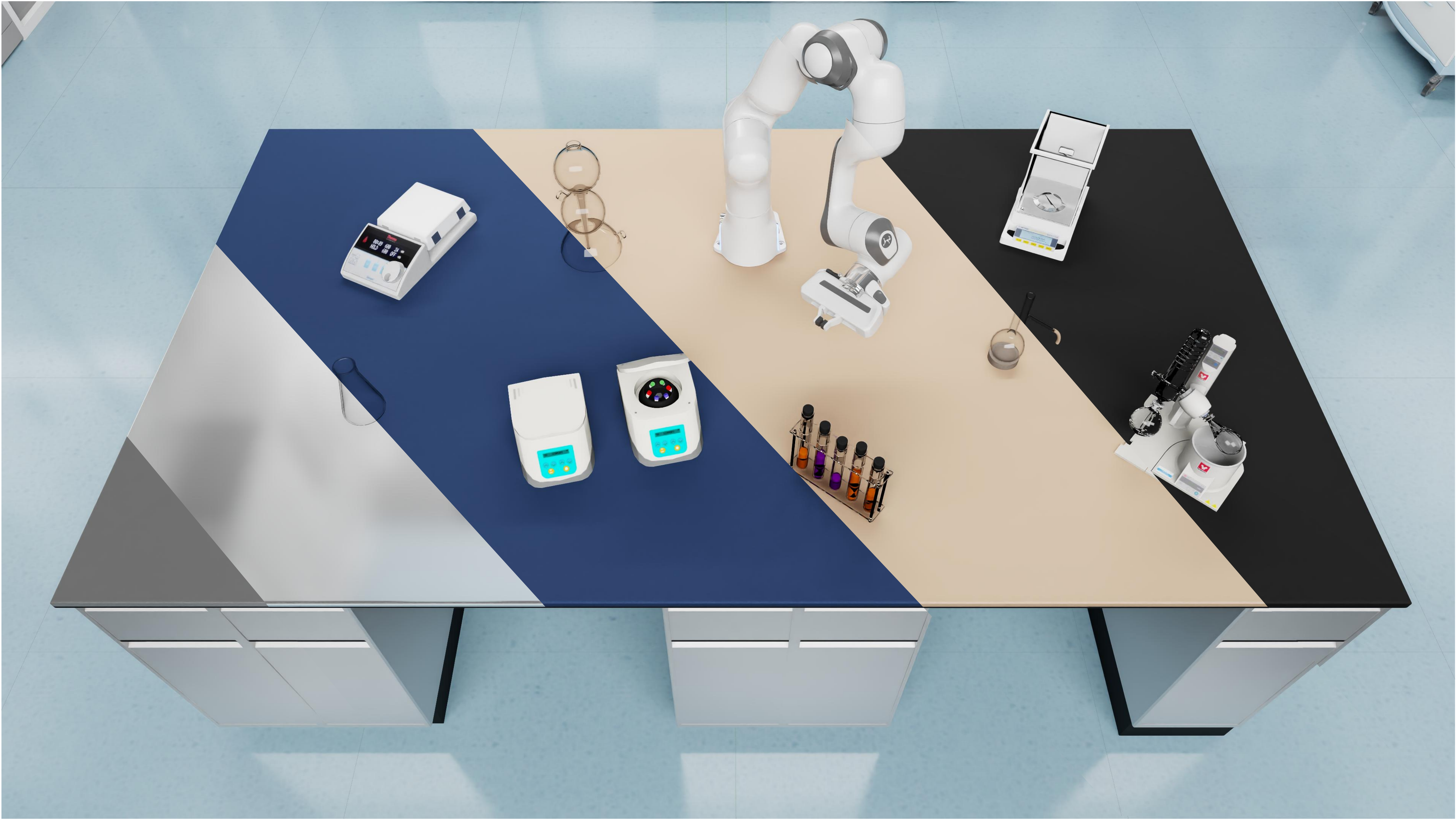}
    \caption{Representative visualizations of the lab-centric data scaling strategy. Each column illustrates variations in object spatial layouts, background distractors, workbench materials, and illumination conditions, while preserving operational feasibility and experimental validity.}
    \label{fig:data_scaling_vis}
\end{figure}

We visualize variations along four dimensions as shown in Figure~\ref{fig:data_scaling_vis}.
First, spatial stochasticity is introduced by perturbing the positions and orientations of target experimental assets within safety-constrained and collision-free regions, resulting in diverse manipulation layouts.
Second, background distractions are incorporated by placing semantically plausible but task-irrelevant laboratory objects (e.g., containers or holders) on the workbench, increasing visual clutter while maintaining physical realism.
Third, material heterogeneity is reflected through randomized workbench textures drawn from a curated library of high-fidelity laboratory surface materials, including both Isaac Sim assets and real-world chemical lab benchmarks.
Finally, illumination variability is achieved by randomizing light source color, intensity, and spatial configuration, producing substantial changes in shading, reflections, and object appearance.

Together, these visual variations substantially expand the training distribution and expose the policy to a wide range of realistic laboratory conditions, thereby improving robustness and generalization in sim-to-real trajectory execution.

\section{Protocols and Task Grouping}
\label{app:Protocols}

In this section, we detail the protocols for each task level, providing a structured approach to assess the performance of embodied agents across a variety of tasks. These protocols are designed to represent different levels of task complexity, ranging from short-horizon tasks to ultra-long-horizon tasks.

\subsection{Level 1: Short-Horizon Tasks with Single Target}
\label{app:level1}
Level 1 tasks consist of 3-step interactions focused on a single manipulation target. These tasks are designed to evaluate the agent's precision and efficiency in executing simple, short-duration actions. Examples include pouring a reagent into a container and returning it to its original position, or activating a heating plate after placing a beaker. The agent is required to complete a sequence of actions with high accuracy and minimal error. Representative examples of the detailed protocols for this level are provided below:

\begin{tcolorbox}[
  enhanced,
  colback=white,
  colframe=black,
  colbacktitle=black,
  fonttitle=\bfseries\itshape\color{white},
  title={Protocol 1: Heat Plate Placement},
  arc=3pt,
  boxrule=1pt,
  breakable
]

\textbf{Problem:} 

Place dichloromethane (5.0 g, 0.041 mol) from the narrow-mouth bottle on the heating plate. Press the heating plate button to start heating. Place the narrow-mouth bottle back in its position.

\vspace{0.8em}
\noindent\colorbox{blue!10}{
\parbox{\dimexpr\linewidth-2\fboxsep}{
\textbf{Assets Used:} Heating Plate, Narrow-Mouth Bottle (Liquid)
}}

\vspace{0.2em}
\noindent\colorbox{gray!10}{
\parbox{\dimexpr\linewidth-2\fboxsep}{
\textbf{Atomic Actions Sequence} Pick-Place-Press
}}

\end{tcolorbox}


\begin{tcolorbox}[
  enhanced,
  colback=white,
  colframe=black,
  colbacktitle=black,
  fonttitle=\bfseries\itshape\color{white},
  title={Protocol 2: Weigh p-Toluidine},
  arc=3pt,
  boxrule=1pt,
  breakable
]

\textbf{Problem:} 

Use an electronic balance to weigh p-toluidine (5.0 g, 0.041 mol) from the wide-mouth bottle. 

Place the weighed p-toluidine back in the original position. 

Pick up the weighing container for the next step.

\vspace{0.8em}
\noindent\colorbox{blue!10}{
\parbox{\dimexpr\linewidth-2\fboxsep}{
\textbf{Assets Used:} Electronic Scale, Wide-Mouth Bottle (Solid)
}}

\vspace{0.2em}
\noindent\colorbox{gray!10}{
\parbox{\dimexpr\linewidth-2\fboxsep}{
\textbf{Atomic Actions Sequence} Pick-Place-Press
}}

\end{tcolorbox}

\subsection{Level 2: Long-Horizon Tasks with Multiple Targets}
\label{app:level2}
Level 2 tasks involve 7 atomic actions and 3-4 manipulation targets. These tasks are more complex than Level 1, requiring longer action sequences and the coordination of multiple manipulation objects. In addition to precise execution, the agent must demonstrate reasoning and planning capabilities to handle task complexity and adapt to changes in the manipulation sequence or targets. Representative examples of the detailed protocols for this level are provided below:

\begin{tcolorbox}[
  enhanced,
  colback=white,
  colframe=black,
  colbacktitle=black,
  fonttitle=\bfseries\itshape\color{white},
  title={Protocol 10: Stir and Add Sulfuric Acid},
  arc=3pt,
  boxrule=1pt,
  breakable
]

\textbf{Problem:} 

Place benzoic acid (5.0 g, 0.041 mol) from the wide-mouth bottle on the heating plate and press the button to start stirring until the benzoic acid partially dissolves.

Slowly pour concentrated sulfuric acid (0.5 mL) from an Erlenmeyer flask into the benzoic acid while stirring to avoid local overheating.

Place the Erlenmeyer flask containing concentrated sulfuric acid back in its original position.

\vspace{0.8em}
\noindent\colorbox{blue!10}{
\parbox{\dimexpr\linewidth-2\fboxsep}{
\textbf{Assets Used:} Erlenmeyer Flask, Heating Plate, Narrow-Mouth Bottle (Liquid)
}}

\vspace{0.2em}
\noindent\colorbox{gray!10}{
\parbox{\dimexpr\linewidth-2\fboxsep}{
\textbf{Atomic Actions Sequence} Pick-Place-Press-Pick-Pour-Place-Press
}}

\end{tcolorbox}


\begin{tcolorbox}[
  enhanced,
  colback=white,
  colframe=black,
  colbacktitle=black,
  fonttitle=\bfseries\itshape\color{white},
  title={Protocol 11: Add Methanol after Weighing},
  arc=3pt,
  boxrule=1pt,
  breakable
]

\textbf{Problem:} 

Use an electronic balance to weigh benzoic acid (5.0 g, 0.041 mol) from the wide-mouth bottle.

Add methanol (20 mL, 0.49 mol) from the narrow-mouth bottle to the benzoic acid.

Place the containers back after use.

Pick up the round-bottom flask for the next operation.

\vspace{0.8em}
\noindent\colorbox{blue!10}{
\parbox{\dimexpr\linewidth-2\fboxsep}{
\textbf{Assets Used:} Electronic Scale, Erlenmeyer Flask, Wide-Mouth Bottle (Solid), Narrow-Mouth Bottle (Liquid)
}}

\vspace{0.2em}
\noindent\colorbox{gray!10}{
\parbox{\dimexpr\linewidth-2\fboxsep}{
\textbf{Atomic Actions Sequence} Pick-Place-Press-Pick-Pour-Place-Press
}}

\end{tcolorbox}

\subsection{Level 3: Ultra-Long-Horizon Tasks} 
\label{app:level3}
Level 3 tasks are the most complex, involving 11 atomic actions and over 4 manipulation targets. These tasks require the agent to possess strong memory, long-term planning, and error-correction abilities to manage the extended sequence of actions. The agent must maintain consistency and coherence across the entire task execution, ensuring that the goals are achieved despite the increased complexity and potential for error accumulation. Representative examples of the detailed protocols for this level are provided below:

\begin{tcolorbox}[
  enhanced,
  colback=white,
  colframe=black,
  colbacktitle=black,
  fonttitle=\bfseries\itshape\color{white},
  title={Protocol 18: Esterification Charge Sequence},
  arc=3pt,
  boxrule=1pt,
  breakable
]

\textbf{Problem:} 

Use an electronic balance to weigh benzoic acid (5.0 g, 0.041 mol) from the wide-mouth bottle, then place the Erlenmeyer flask containing benzoic acid on the heating plate, press the button to start stirring, add methanol (20 mL, 0.49 mol) from the narrow-mouth bottle to the benzoic acid, and finally slowly pour concentrated sulfuric acid (0.5 mL, catalyst and dehydrating agent) into the benzoic acid.

\vspace{0.8em}
\noindent\colorbox{blue!10}{
\parbox{\dimexpr\linewidth-2\fboxsep}{
\textbf{Assets Used:} Electronic Scale, Erlenmeyer Flask, Heating Plate, Wide-Mouth Bottle (Solid), Narrow-Mouth Bottle (Liquid)
}}

\vspace{0.2em}
\noindent\colorbox{gray!10}{
\parbox{\dimexpr\linewidth-2\fboxsep}{
\textbf{Atomic Actions Sequence} Pick-Weigh-Pick-Place-Pick-Pour-Place-Press-Shake-Pick-Place
}}

\end{tcolorbox}

\section{Additional Sim-to-Real Results}
\label{app:Sim2real}

\begin{figure}[H]
    \centering
    \includegraphics[width=\linewidth]{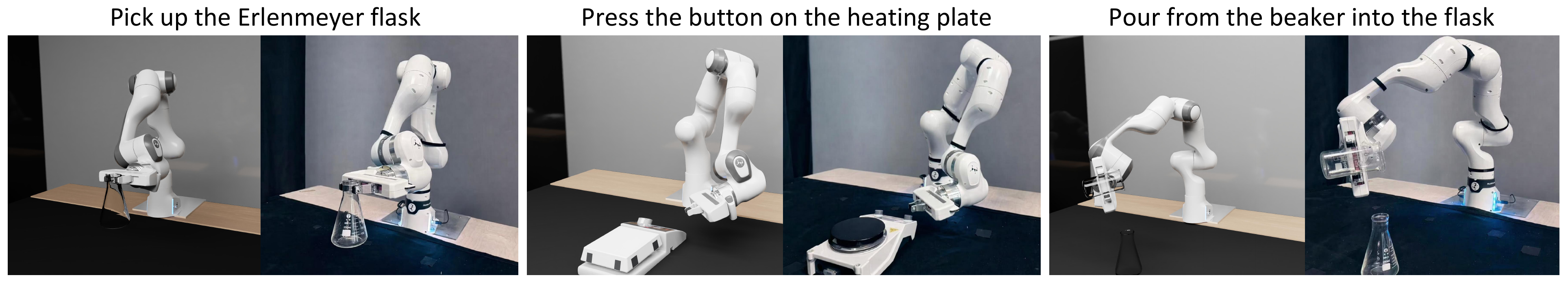}
    \caption{Visual comparison of sim-to-real transfer. For two representative manipulation tasks, we compare the simulated environment (left) with the corresponding real-world execution (right), showing consistent object placement, robot configurations, and task progress}
    \label{fig:sim2real}
\end{figure}

To evaluate the usability and robustness of the generated expert trajectories in real-world settings, we conducted trajectory replay experiments on a physical robot using the collected data. Figure~\ref{fig:sim2real} illustrates the visual appearance of both the simulator and the real environment. Specifically, under randomized configurations, we sampled 300 trajectories from 20 chemical experiment protocols across three difficulty levels. For each trajectory, we reconstructed the object placement scene in the physical environment to match the configuration used during data collection and replayed the complete action sequence, thereby assessing the continuity and execution accuracy of the policy in real-world conditions. We further investigated the impact of incorporating a reset-based correction mechanism during data collection on real-robot trajectory replay performance. In this evaluation, a trajectory replay was considered successful if the robot completed the entire action sequence without task failure, such as object displacement errors or execution interruptions.

\begin{figure}[H]
\centering
\includegraphics[width=0.45\linewidth]{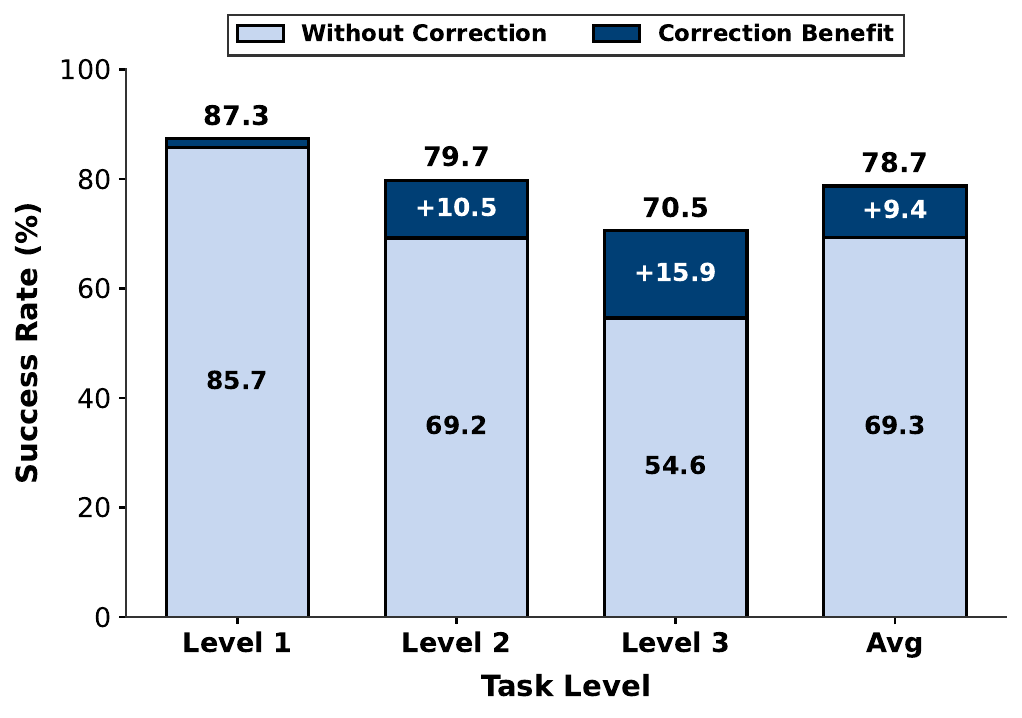}
\caption{Sim-to-real evaluation of open-loop trajectory replay on a physical robot. Success rates are reported over 100 trajectories per difficulty level, comparing cases with and without reset correction actions inserted between pick-and-place executions.}
\label{fig:sim_to_real}
\end{figure}

As shown in Figure~\ref{fig:sim_to_real}, without the reset mechanism, the success rates for all three levels exceeded 50\%, with Level~1 achieving 85.7\%, Level~2 69.2\%, and Level~3 54.6\%. This indicates that the expert trajectories possess high quality and dynamic feasibility. At the same time, the rapid decline in success rates with increasing task difficulty highlights the pronounced accumulation of errors in long-horizon tasks.

To quantitatively assess the benefits of the reset-based correction mechanism, we compared trajectory replay success rates with and without the insertion of reset actions. Specifically, a reset action was applied after each pick-and-place action pair, returning the robotic arm to an initial configuration well within joint limits to facilitate subsequent execution.

As illustrated in Figure~\ref{fig:sim_to_real}, introducing reset actions consistently improved trajectory replay performance across all difficulty levels. For short-horizon tasks (Level~1), the success rate increased slightly from 85.7\% to 87.3\%, reflecting minimal error accumulation at this stage. In contrast, for longer and more complex tasks, the advantage of the reset mechanism became more pronounced: Level~2 success rates improved from 69.2\% to 79.7\%, and Level~3 from 54.6\% to 70.5\%. Overall, the reset-based correction mechanism yielded an average absolute improvement of 9.4\% across all tasks (from 69.3\% to 78.7\%).

These results demonstrate that (i) the expert trajectories exhibit high quality and dynamic feasibility, and (ii) the reset-based correction mechanism plays a critical role in mitigating error accumulation and maintaining stable and accurate execution for long-horizon, fine-grained manipulation tasks in real-world trajectory replay settings.

\section{Policy Training and Inference Details}
\label{app:Policy}

\subsection{Policy Architectures}

We benchmark three commonly used policy families for imitation learning in robotic manipulation: ACT, Diffusion Policy (DP), and $\pi_0$. These approaches differ in how they incorporate temporal context and construct action sequences.

\textbf{ACT} employs an autoregressive transformer to generate low-level control trajectories from visual observations and robot proprioception. Its predictions are generated sequentially, with each action conditioned on previously observed states and earlier actions. This design enables the policy to produce and execute actions in temporal chunks.

\textbf{Diffusion Policy (DP)} represents control generation as a conditional denoising problem. Based on the available observation history, it starts from an initial noise sample and progressively removes the noise to obtain a sequence of future actions.

\textbf{$\pi_0$} adopts a vision-language-action architecture built on the PaliGemma backbone. A separate action expert, trained with flow matching, converts the learned representation into continuous control commands. This formulation is designed to handle manipulation behaviors with extended temporal horizons and compositional task structures.

\subsection{Model Inputs and Outputs}

To make the comparison consistent, we expose all policies through the same observation and control interface. At every decision point, each policy receives visual information from the workspace together with the robot's proprioceptive state, and produces continuous commands in joint space.

The visual input contains RGB images with a resolution of $256 \times 256$. Three camera viewpoints are available: a wrist-mounted camera, a front-facing camera directed toward the workspace, and an overhead camera providing a global view of the scene. ACT and Diffusion Policy use the front-facing and overhead images, whereas $\pi_0$ uses the wrist-mounted and front-facing views to provide a more egocentric perspective. The absolute joint positions of the robot are supplied to all three policies. Given these inputs, the policy predicts the target joint configuration for the next control step, which is then executed in a closed-loop controller.

\subsection{Training Details}

We follow the standard training recipes associated with the official or open-source implementations of the three policies. We modify these implementations only where necessary to accommodate task-specific settings and the shared observation-action interface. ACT and Diffusion Policy are trained in PyTorch for 200 epochs on a single NVIDIA RTX 4090 GPU. Both models use AdamW with a fixed learning rate of $1 \times 10^{-4}$, while an exponential moving average (EMA) of the parameters is updated during training. The batch size is 128 for ACT and 64 for Diffusion Policy.

The $\pi_0$ policy is implemented in JAX and optimized for $15{,}000$ update steps on four NVIDIA H200 GPUs. Its learning rate is also fixed at $1 \times 10^{-4}$. The policies receive different amounts of temporal context: ACT and $\pi_0$ make predictions from one observation frame, whereas Diffusion Policy consumes three consecutive frames. The action chunk length is set to 60 for ACT and Diffusion Policy, and to 8 for $\pi_0$. For every policy, the input modalities are limited to RGB images and robot joint positions. Unless stated otherwise, we use the default hyperparameter settings provided by the corresponding original implementations.

\section{Prompts Used for Data Generation and Evaluation}
\label{app:prompts}
This appendix provides detailed descriptions of the various prompts used in the generation and evaluation of data for robotic tasks, chemical experimental protocols, and controller debugging. Each prompt serves a specific function in the task pipeline, ensuring that data is structured correctly and that system behavior is optimized.

\subsection{Prompt for Scene and Action Extraction}
\label{app:prompt_protocol}

The \textbf{Protocol-to-Action Parsing} prompt is designed to convert natural-language laboratory protocols into structured representations. The output consists of two main components: an equipment inventory and a sequence of atomic actions. This helps standardize complex laboratory procedures into executable instructions for robotic systems.

\textbf{Key Features:} This prompt uses a closed asset vocabulary to map any equipment mentioned in the protocol to a predefined set of interactive containers (e.g., Beakers, Flasks) and static devices (e.g., Heating Plates, Scales). Static devices are activated using specific controllers like \textit{PressController} for horizontal activation. Additionally, liquid and solid reagents are assigned to specific containers, ensuring proper handling in the robotic system. It also enforces action preconditions for controllers like \textit{PourController} and \textit{ShakeController}, which require the object to be held by the gripper beforehand.

The specific prompt is as follows:
\vspace{1em}
\begin{promptbox}[Prompt for Scene and Action Extraction]
You are an assistant for converting natural-language laboratory protocols into structured representations for execution by embodied agents.\\

You must extract two components:\\
1. \textbf{Equipment Inventory:} list all relevant laboratory assets.\\
2. \textbf{Atomic Action Sequence:} decompose the protocol into ordered atomic actions.\\

Special constraints and rules:\\
1. \textbf{Asset Mapping:} Map all objects to a predefined set of containers and devices. Unlisted items should be approximated to similar valid assets.\\
2. \textbf{Device Operation:} Some devices require activation via button pressing; others do not.\\
3. \textbf{Reagent Abstraction:} Distinguish only between solid and liquid reagents, assigning them to representative containers.\\
4. \textbf{Placement Handling:} Use a generic placement target when returning objects to unspecified locations.\\
5. \textbf{Atomic Actions:} Use a fixed set of primitive controllers (e.g., pick, place, pour, open, close, press, shake).\\
6. \textbf{Action Dependency:} Certain actions require prior grasping without interruption.\\

Output format requirement:\\
You must return exactly two blocks enclosed by the following markers:\\

\begin{verbatim}
===EQUIPMENT_START===
...
===EQUIPMENT_END===

===ACTIONS_START===
...
===ACTIONS_END===
\end{verbatim}

Final instruction:\\
Ensure consistency between the inventory and action sequence. The generated program should be complete and executable under the defined constraints.\\

Protocol: \{input\_protocol\} \\

Structured Output:
\end{promptbox}

\subsection{Prompt for Controller-Class Code Generation}
\label{app:prompt_controller_code}

The \textbf{Controller-Class Code Generation} prompt focuses on generating task-specific robotic controller classes. The generated code must adhere to the project framework and support dual-mode operation (execution and data acquisition). The controller must be compatible with predefined actions like \textit{Pick}, \textit{Pour}, \textit{Move}, and \textit{Shake}.

\textbf{Key Features:} This prompt ensures dual-mode support, allowing the controller to operate in both "infer" (execution) and "collect" (data acquisition) modes. In "infer" mode, the controller executes predefined actions step by step, interacting with objects and completing tasks as specified. In "collect" mode, the controller is responsible for gathering relevant data, such as sensor readings, visual data from cameras, and other measurements that help analyze the performance of the robotic system. The controller must inherit from the \textit{BaseController} class, ensuring that it follows the established project framework, which includes automatic initialization of key components such as the motion planning controller. It must also implement essential methods, such as \textit{reset()} to reset the controller’s state between task executions and \textit{step(state)} to advance the controller’s operation based on the current system state. Case sensitivity plays a critical role in ensuring that object names are used exactly as specified in the program, such as \textit{"Beaker1"} and \textit{"ElectronicScale"}. This prevents errors due to miscapitalization, which could otherwise cause mismatches in state lookups or action assignments. Furthermore, the controller is designed to handle missing or incomplete state information gracefully. If any state variable is absent or undefined, the controller returns a default response of \textit{(None, False, False)} to maintain stability and prevent crashes. These design choices ensure that the controller operates efficiently, smoothly handles missing or inconsistent data, and remains robust against runtime errors during task execution.

The specific prompt is as follows:

\vspace{1em}
\begin{promptbox}[Prompt for Controller-Class Code Generation]
You are an assistant for generating task-specific robotic controller code following a predefined framework, supporting both execution and data collection modes.\\

\textbf{Core requirements:}\\
1. \textbf{Class structure:} Inherit from a base controller and implement \textit{reset()} and \textit{step(state)}.\\
2. \textbf{Dual-mode support:} The controller operates under different modes, with data collection enabled only in a specific mode and guarded by conditional checks.\\
3. \textbf{Naming consistency:} Object and state identifiers must remain consistent with the given task specification.\\
4. \textbf{Safe access:} Missing state entries should not cause errors; instead, return a default safe output and wait for the next step.\\
5. \textbf{Monitoring hooks:} Track current targets and action types during execution.\\
6. \textbf{Controller usage:} Use the provided base controller utilities without redefining low-level components.\\

\textbf{Atomic action constraints:}\\
Only predefined primitive controllers may be used, following standard constructor patterns and shared control interfaces.\\

\textbf{Execution conventions:}\\
Use a default orientation and apply task-dependent parameter adjustments where necessary.\\

\textbf{Grasp maintenance:}\\
Maintain object state consistency when an object is held during execution.\\

\textbf{Orchestration:}\\
Organize actions via staged execution (e.g., finite-state transitions), updating state and returning appropriate signals at each step.\\

\textbf{Data collection:}\\
When enabled, cache relevant step data under guarded conditions, and store final state information before termination.\\

\textbf{Output format:}\\
Return only valid Python code for a complete controller class, without additional explanations.\\

Task specification: \{task\_instruction\}\\
Generated controller code:
\end{promptbox}

\subsection{Prompt for Controller-Parameter Optimization}
\label{app:prompt_controller_param}

The \textbf{Controller-Parameter Optimization} prompt assists in debugging and optimizing robotic controller parameters by analyzing execution logs. It helps identify failure causes and recommends targeted parameter adjustments to improve the success rate without altering the controller's logic.

\textbf{Key Features:} This prompt requires minimal modifications to the code structure, focusing only on parameter values to optimize performance. Changes are driven by analysis of execution logs, not trial-and-error methods. The prompt also emphasizes memory-based reuse, where previous failures with similar conditions are used to inform parameter adjustments. All modifications must maintain the safety and compatibility of the controller, ensuring it remains executable and free from errors.

The specific prompt is as follows:

\vspace{1em}
\begin{promptbox}[Prompt for Controller-Parameter Optimization]
You are an assistant for diagnosing execution failures and improving controller parameters based on logs and code. Your goal is to identify likely causes and apply minimal parameter adjustments without altering the controller structure.\\

\textbf{Goal:} Given an execution log and controller code, determine which parameter(s) to adjust and update the code to improve performance while preserving the original logic.\\

\textbf{Key constraints:}\\
1. \textbf{Structure preservation:} Keep class structure, methods, and control flow unchanged; only modify parameter values.\\
2. \textbf{Evidence-based edits:} Use log signals to infer necessary adjustments; avoid arbitrary or exhaustive search.\\
3. \textbf{Minimal changes:} Prefer small, targeted updates linked to the failure.\\
4. \textbf{Compatibility:} Ensure the updated code remains safe and executable.\\

\textbf{Experience-based reuse:}\\
Consider similar past cases based on action type, failure type, or general constraints. If applicable, briefly summarize prior adjustments and decide whether to reuse them; otherwise proceed with independent reasoning.\\

\textbf{Decision procedure:}\\
Step 1: Extract key failure signals from the log.\\
Step 2: Identify the likely cause category (e.g., alignment, collision, reachability, interaction mismatch).\\
Step 3: Select relevant parameter adjustments and justify the choice.\\
Step 4: Apply edits while keeping all other code unchanged.\\

\textbf{Output format (strict):}\\
Your output must contain exactly two parts separated by \texttt{===ANALYSIS\_END===}.\\
(1) \textbf{Analysis and modification plan}:\\
\hspace*{1em}- brief diagnosis\\
\hspace*{1em}- reuse decision (if any)\\
\hspace*{1em}- parameter changes and rationale\\
(2) \textbf{Updated Python code}:\\
\hspace*{1em}- same structure, only parameters edited\\

Execution log: \{exec\_log\}\\
Original controller code: \{controller\_code\}\\
Optimized output:
\end{promptbox}

\subsection{Prompt for Layout Optimization}
\label{app:prompt_layout}

The \textbf{Prompt for Layout Optimization} is designed to optimize the placement of objects in a scene, particularly when a gripper fails to manipulate a target object. The prompt analyzes execution logs, identifies unreachable objects, and makes minimal adjustments to object positions to facilitate successful operation.

\textbf{Key Features:} This prompt ensures that all objects are placed at a fixed height of 0.775 to maintain consistency in object handling. It also ensures that objects must be within an elliptical reachability region for the gripper to access them, with a specific focus on optimizing placement to improve gripper success. The prompt works to prevent collisions by ensuring objects' bounding boxes do not overlap and reducing occlusion, particularly along the x-axis. Additionally, it searches past cases for similar failures and reuses successful strategies where applicable.

The specific prompt is as follows:

\begin{promptbox}[Prompt for Layout Optimization]
You are an assistant for optimizing scene layouts when repeated action-level adjustments fail. Your goal is to improve object placement to enable successful interaction.\\

\textbf{Task:}\\
1. \textbf{Analyze Logs:} Identify failure patterns and relative object--gripper relationships.\\
2. \textbf{Identify Target Object:} Focus on the object involved in the last failed action.\\
3. \textbf{Single-Object Adjustment:} Modify only one object's position at a time.\\

\textbf{Constraints:}\\
\textbf{Height:} All objects share a fixed placement height.\\
\textbf{Reachability:} Objects must lie within a predefined reachable region.\\
\textbf{Collision:} Avoid overlap between objects.\\
\textbf{Occlusion:} Reduce blocking effects between objects.\\
\textbf{Modification Scope:} Only position attributes may be changed; other fields must remain unchanged.\\
\textbf{Semantics:} Maintain basic spatial relationships where applicable.\\

\textbf{Experience-based retrieval:}\\
Before adjustment, consider similar past cases based on object type, failure type, or spatial constraints, and reuse effective strategies when appropriate.\\

\textbf{Optimization process:}\\
1. Extract key failure signals from logs.\\
2. Select the problematic object.\\
3. Adjust its position under constraints.\\
4. Verify that constraints are satisfied.\\

\textbf{Guidelines:}\\
Prefer small adjustments that improve reachability and reduce conflicts, without introducing new issues.\\

\textbf{Input:}\\
Execution logs and current scene representation.\\

\textbf{Output format:}\\
Return the updated scene JSON with only one object's position modified. Do not include explanations.\\

\begin{verbatim}
{
  "objects": [
    {
      "name": "Object1",
      "type": "Type1",
      "position": [x1, y1, z],
      ...
    },
    {
      "name": "Object2",
      "type": "Type2",
      "position": [x2, y2, z],
      ...
    }
  ]
}
\end{verbatim}

Generated output:

\end{promptbox}

\section{Structured Monitoring Feedback and Memory Representations}
\label{app:monitoring}

This section presents a concrete, end-to-end example illustrating how structured monitoring
feedback drives refinement decisions and how the resulting signals are represented and
consumed by downstream modules.
Rather than reporting isolated failures, we trace a single rollout in which the monitor
progressively diagnoses execution-level errors, applies local action refinement, and
ultimately escalates to scene refinement when execution-level corrections prove insufficient.

All logs shown below are lightly trimmed for readability, while preserving the key numerical
signals and decision points used by the monitor.
Together, the four blocks form a coherent failure–refinement trace, demonstrating how
monitor outputs serve as both (i) immediate supervision signals for refinement and
(ii) structured memory entries for future task synthesis.


\begin{tcolorbox}[
  enhanced,
  colback=white,
  colframe=black,
  colbacktitle=black,
  fonttitle=\bfseries\itshape\color{white},
  title={Monitor Output: Failure Detection and Attribution},
  arc=3pt,
  boxrule=1pt,
  breakable
]
\textbf{Context.}
The monitor is initialized with a controller and infers the current action as \texttt{pick}.
It tracks the end-effector and target grasp pose online, followed by a verification step for \texttt{Stage 0 -- pick}.

\textbf{Key log evidence (trimmed).}
\vspace{-2pt}
\footnotesize
\begin{verbatim}
=== Monitor Initialized ===
[Monitor] Action type: pick
[Stage 0 - pick] START
...
[Verify Pick] Object: target_object
[Verify Pick] Gripper: (x1, y1, z1)
[Verify Pick] Target: (x2, y2, z2)
[Verify Pick] Result: Distance ~0.37m (Expected <0.1m)
[Stage 0 - pick] FAILED - grasp_failed
!!! TERMINATION - Stage 0 failed
ACTION_FAILED=1
\end{verbatim}
\normalsize
\vspace{-4pt}

\textbf{Diagnosis.}
The failure occurs because the end-effector does not reach the intended grasp region.
At verification, the distance between the gripper and target is approximately $0.37$\,m,
which significantly exceeds the success threshold ($<0.1$\,m).
This suggests a large pose or approach deviation rather than a minor execution error.
\end{tcolorbox}

\begin{tcolorbox}[
  enhanced,
  colback=white,
  colframe=black,
  colbacktitle=black,
  fonttitle=\bfseries\itshape\color{white},
  title={Action Refinement Output: Execution-Level Correction},
  arc=3pt,
  boxrule=1pt,
  breakable
]
\textbf{Trigger.}
The monitor reports a \texttt{grasp\_failed} event with a large deviation between the end-effector and target,
indicating an execution-level mismatch during the \texttt{pick} action.

\textbf{Key refinement output (trimmed).}
\vspace{-2pt}
\footnotesize
\begin{verbatim}
[ActionRefine] Stage: pick
[ActionRefine] Error: grasp_failed
[ActionRefine] Target: object
[ActionRefine] Param: orientation
[ActionRefine] Candidates: [...]
[ActionRefine] Update applied
\end{verbatim}
\normalsize
\vspace{-4pt}

\textbf{Interpretation.}
A local adjustment is applied to the end-effector orientation while keeping other factors unchanged,
aiming to improve alignment without modifying the overall task structure.

\textbf{Expected effect.}
If the failure is caused by misalignment, this update is expected to reduce the deviation
and enable successful execution.
\end{tcolorbox}

\paragraph{From execution-level correction to scene-level diagnosis.}
While action refinement introduces targeted execution-level adjustments, its effectiveness is
continuously evaluated by the monitor using the same geometric verification criteria.
As shown next, the monitor explicitly reasons over progress trends across multiple refinement
iterations and escalates to scene refinement only when improvement stalls, avoiding premature
or unnecessary scene perturbations.


\begin{tcolorbox}[
  enhanced,
  colback=white,
  colframe=black,
  colbacktitle=black,
  fonttitle=\bfseries\itshape\color{white},
  title={Monitor Output: Escalation after Repeated Action Refinement Failures},
  arc=3pt,
  boxrule=1pt,
  breakable
]
\textbf{Context.}
After an initial failure, the system applies iterative action refinement while monitoring
the end-effector--target deviation and progress across iterations.

\textbf{Key log evidence (trimmed).}
\vspace{-2pt}
\footnotesize
\begin{verbatim}
[Monitor] Refinement iteration: 1
[Verify] Deviation: high (expected: low)
[Monitor] No sufficient progress

[Monitor] Refinement iteration: 2
[Verify] Deviation: high (expected: low)
[Monitor] No sufficient progress

[Monitor] Refinement iteration: 3
[Verify] Deviation: high
[Monitor] Progress stalled
[Monitor] Escalation condition met
[Monitor] Escalate to Scene Refinement
\end{verbatim}
\normalsize
\vspace{-4pt}

\textbf{Diagnosis.}
Despite multiple refinement attempts, the deviation remains high and shows no consistent improvement.
The monitor therefore attributes the failure to scene-level infeasibility rather than local execution
errors, and triggers escalation.
\end{tcolorbox}

\begin{tcolorbox}[
  enhanced,
  colback=white,
  colframe=black,
  colbacktitle=black,
  fonttitle=\bfseries\itshape\color{white},
  title={Scene Refinement Output: Layout and Anchor Parameter Update},
  arc=3pt,
  boxrule=1pt,
  breakable
]
\textbf{Trigger.}
Scene refinement is triggered after repeated action refinements fail to reduce the deviation,
indicating a layout-level issue.

\textbf{Key refinement output (trimmed).}
\vspace{-2pt}
\footnotesize
\begin{verbatim}
[SceneRefine] Triggered by monitor
[SceneRefine] Stage: pick
[SceneRefine] Target: object

[SceneRefine] Detected issues:
  - grasp region not reachable
  - partial occlusion in approach path

[SceneRefine] Updates:
  - adjust object position
  - modify object orientation
  - update grasp anchor

[SceneRefine] Post-check: constraints satisfied
[SceneRefine] Scene update committed
\end{verbatim}
\normalsize
\vspace{-4pt}

\textbf{Interpretation.}
Scene refinement applies structured adjustments to object layout and grasp configuration
to resolve geometric constraints and improve feasibility.

\textbf{Expected effect.}
After refinement, reachability and safety conditions are satisfied, enabling successful execution
in subsequent attempts without further scene-level intervention.
\end{tcolorbox}

\paragraph{Implications for structured memory and reuse.}
Importantly, both the monitor diagnoses and the applied scene updates are stored as structured
memory entries, including numeric pose deltas, anchor offsets, and violated constraints.
These records allow future task instantiations with similar layouts or protocols to bypass
repeated trial-and-error refinement, directly initializing scenes in a feasible region.
This example illustrates how monitoring feedback in RoboChemGym not only resolves individual
failures, but also accumulates reusable knowledge across episodes.


\end{document}